\documentclass[journal]{IEEEtran}
\usepackage{cite}
\usepackage{amsmath,amssymb,amsfonts}
\usepackage{algorithmic}
\usepackage{graphicx}
\usepackage{textcomp}
\usepackage{booktabs}
\usepackage{wrapfig}
\usepackage{bbding}
\usepackage{threeparttable}
\usepackage{multirow}
\usepackage{hhline}

\usepackage{algorithm}
\usepackage{array}
\usepackage[caption=false,font=normalsize,labelfont=sf,textfont=sf]{subfig}
\usepackage{stfloats}
\usepackage{url}
\usepackage{verbatim}
\usepackage{cite}
\begin{document}
\title{Robust Video-Based Pothole Detection and Area Estimation for Intelligent Vehicles with Depth Map and Kalman Smoothing}
\author{Dehao Wang, Haohang Zhu, Yiwen Xu, and Kaiqi Liu
\thanks{It is sponsored by State Key Laboratory of Intelligent Green Vehicle and Mobility under Project No. KFY2416. (\textit{Corresponding Author: Kaiqi Liu})}
\thanks{Dehao Wang is with the School of Integrated Circuits and Electronics, Beijing Institute of Technology, Beijing 100081, China (e-mail: wonderhow@bit.edu.cn).}
\thanks{Haohang Zhu, Yiwen Xu and Kaiqi Liu are with the School of Information and Electronics, Beijing Institute of Technology, Beijing 100081, China (e-mail: 1120221140@bit.edu.cn; 1120221068@bit.edu.cn; liukaiqi@bit.edu.cn).}}

\markboth{Journal of \LaTeX\ Class Files,~Vol.~14, No.~8, August~2021}%
{Shell \MakeLowercase{\textit{et al.}}: A Sample Article Using IEEEtran.cls for IEEE Journals}


\maketitle

\begin{abstract}
Road potholes pose a serious threat to driving safety and comfort, making their detection and assessment a critical task in fields such as autonomous driving. When driving vehicles, the operators usually avoid large potholes and approach smaller ones at reduced speeds to ensure safety. Therefore, accurately estimating pothole area is of vital importance. Most existing vision-based methods rely on distance priors to construct geometric models. However, their performance is susceptible to variations in camera angles and typically relies on the assumption of a flat road surface,  potentially leading to significant errors in complex real-world environments. To address these problems, a robust pothole area estimation framework that integrates object detection and monocular depth estimation in a video stream is proposed in this paper. First, to enhance pothole feature extraction and improve the detection of small potholes, ACSH-YOLOv8 is proposed with ACmix module and the small object detection head. Then, the BoT-SORT algorithm is utilized for pothole tracking, while DepthAnything V2 generates depth maps for each frame. With the obtained depth maps and potholes labels, a novel Minimum Bounding Triangulated Pixel (MBTP) method is proposed for pothole area estimation. Finally, Kalman Filter based on Confidence and Distance (CDKF) is developed to maintain consistency of estimation results across consecutive frames. The results show that ACSH-YOLOv8 model achieves an AP(50) of 76.6\%, representing a 7.6\% improvement over YOLOv8. Through CDKF optimization across consecutive frames, pothole predictions become more robust, thereby enhancing the method's practical applicability.

\end{abstract}


\maketitle

\section{Introduction}
\label{sec:introduction}
\IEEEPARstart{A}{utonomous} driving technology has developed rapidly in recent years, with the enhancement of road safety becoming one of its key objectives. To ensure driving safety and improve ride comfort, autonomous driving systems must accurately perceive and interpret road conditions. Potholes represent a prevalent type of road surface damage, which may adversely affect driving safety\cite{10401010}. Their formation is influenced by a complex array of factors, including natural elements such as climate change and soil composition\cite{bukhari2023review}, as well as human-induced factors such as improper road design, inadequate maintenance, and excessive traffic load\cite{muller2016pothole}. Therefore, it is difficult to predict where potholes will appear on roads.\par

Pothole is one of the major causes of traffic accidents. According to the British Automobile Association, 631,852 pothole-related accidents were reported in 2022, marking a five-year high\cite{bbc_science_2023}. Similarly, the Chicago Sun-Times reported 3,597 traffic accidents caused by potholes in the first two months of 2018 alone\cite{Snyder2018}. In addition to compromising road safety, potholes also negatively impact passenger comfort as vehicles traverse these damaged road surfaces\cite{kirbacs2023effects}. Therefore, real-time detection of road potholes has become a critical area of research\cite{ma2022computer}.\par

While extensive research has focused on detecting and localizing potholes, further estimating their area provides even greater practical value for real-world applications\cite{dhiman2019pothole}. From a safety perspective, the size of a pothole directly influences the choice of obstacle avoidance strategies\cite{wu2019road}. For instance, the vehicle can maintain its course when encountering small potholes, but large potholes might necessitate rerouting or emergency braking to ensure safety. From a comfort perspective, estimating pothole area allows vehicles to determine whether to reduce speed, thereby avoiding severe jolts caused by traversing large potholes at high speeds\cite{loprencipe2017ride}. With the rapid development of connected vehicle technologies\cite{taslimasa2023security}, information about pothole areas can be shared across intelligent traffic systems, enabling other autonomous vehicles to plan routes more efficiently and reduce traffic congestion. Furthermore, analyzing changes in pothole area over time can provide valuable insights into road aging trends, supporting road maintenance and urban planning efforts\cite{baba2023identification}.\par

Research on pothole area estimation generally follows two main approaches. The first involves using LiDAR or similar sensors to obtain 3D point cloud data, from which pothole areas are calculated. The second relies on purely vision-based methods that estimate area using object detection and predefined geometric models. However, the former often suffers from high computational costs, while the latter is highly sensitive to camera angles and struggles to perform well on complex terrains. Moreover, most existing methods process only single-frame data, whereas leveraging video streams for pothole area estimation has the potential to significantly enhance robustness and holds strong promise for practical applications.\par

To improve the accuracy and robustness of pothole area estimation while enhancing processing efficiency and reducing costs using only 2D images, a novel and robust pothole area estimation framework is proposed. The framework is built upon a newly designed pothole detection model, ACSH-YOLOv8, combined with the advanced monocular metric depth estimation network, DepthAnything V2. A novel Minimum Bounding Triangulated Pixel (MBTP) method is introduced to estimate pothole areas with improved reliability. To reduce the impact of factors such as lighting variation and camera motion, the Kalman Filter based on Confidence and Distance (CDKF) algorithm is proposed, which leverages consecutive video frames and adjusts estimations based on detection confidence and distance between pothole and camera.\par

The main contributions of the paper are as follows:\par
\begin{enumerate}
\item In order to achieve high-accuracy and robust pothole area estimation, this paper proposes a novel pothole detection and area estimation framework, where a dedicated MBTP method is introduced as the core module for precise pothole area estimation by integrating pothole regions with the depth map.
\item To enhance the model’s detection accuracy for potholes with varying scale and blur, the ACSH-YOLOv8 model is proposed by adding an additional detection head for small potholes, and incorporating a hybrid attention mechanism, ACmix, in the neck of the architecture to improve detail awareness.
\item To enhance the robustness of area estimation in video streams, the CDKF method is proposed, which refines area estimates based on pothole tracking results, utilizing detection confidence and distance as optimization.
\end{enumerate}

The remainder of the paper is organized as follows. Section II provides a review of related work on pothole detection and pothole area estimation. Section III provides a detailed explanation of the core methodology for pothole area estimation, covering pothole detection, tracking, depth estimation, area calculation, and consecutive frame optimization. Section IV introduces the dataset, evaluation metrics, and experimental setup for both the detection model and area estimation algorithm. Section V presents a quantitative and visual analysis of the results, and the work of this paper is summarized in Section VI.
\section{Related Works}
\subsection{Pothole Detection}
There is considerable research on pothole detection, including both traditional machine learning algorithms and deep learning approaches. Traditional machine learning methods such as Otsu's thresholding\cite{goh2018performance}, spectral clustering\cite{buza2013pothole}, and morphological operations\cite{ouma2017pothole} are used to extract and identify potential pothole regions. While these algorithms have the advantage of lower computational load, their classification performance and robustness are often limited. Some studies employ 3D point cloud data, using surface normal information for pothole geometric modeling\cite{wu2021scale}. However, 3D point cloud data acquisition is often costly and computationally demanding. Nowadays, with the rapid development of deep learning technologies, numerous CNN-based deep learning networks for object detection are proposed, creating significant opportunities for pothole detection development. These methods significantly enhance the accuracy and robustness of pothole detection, and the localization of potholes is obtained precisely\cite{10474576}. Among pothole detection algorithms, the one-stage algorithm You Only Look Once (YOLO)\cite{redmon2016you} gains widespread application due to its high accuracy and real-time processing capabilities. Ukhwah\cite{ukhwah2019asphalt} demonstrates the effectiveness of YOLOv3 and its variants in road pothole detection. Shaghouri\cite{shaghouri2021real} introduces CSPDarknet53 as a backbone based on YOLOv4, achieving a balance between accuracy and speed. Mahalingesh\cite{mahalingesh2024pothole} integrates the YOLOv8 algorithm and deploys it on a Raspberry Pi for hardware testing, highlighting the significant potential of YOLO-based algorithms in practical applications.\par

\subsection{Pothole Area Estimation}
Current research on pothole area estimation falls into two main approaches: one involves obtaining 3D point cloud arrays for area estimation, and the other uses 2D images and image processing techniques. In terms of 3D point cloud analysis, Ravi\cite{ravi2020pothole} utilizes LiDAR to capture road point clouds and applies a vehicle motion mapping model to achieve high-precision pothole area estimation. Chen\cite{chen2024volumetric} employs drones to slice images into 3D point clouds and proposes the UAV-Structure-from-Motion algorithm, which uses motion sensing for pothole area estimation. Although these methods achieve high accuracy, they rely on motion models, which may introduce significant errors in trajectory and speed control if the motion varies greatly. This reduces the accuracy of the estimation and limits their practical applicability. Additionally, using LiDAR or high-resolution images to generate point clouds requires considerable computational resources, making these methods costly and unsuitable for real-time applications.\par

For 2D image processing, most research focuses on proposing new area estimation methods based on object detection algorithms. Heo\cite{heo2023image} uses a pinhole camera model and prior distance equations to estimate pothole areas. Kharel\cite{kharel2021potholes} employs Inverse Perspective Mapping (IPM) to convert camera intrinsic parameters and estimate pothole areas. Chitale\cite{chitale2020pothole} applies distance priors and triangle similarity to estimate pothole areas. However, these methods strongly depend on geometric models and are highly sensitive to the camera's viewing angle, making them less adaptable for widespread use. Furthermore, these methods operate under the assumption of planar road surfaces, whereas potholes predominantly form in complex, non-uniform terrains. Conventional prior models fail to adequately represent these geometric irregularities, resulting in significant estimation inaccuracies.\par

\section{Method}

\begin{figure*}[htbp]
\centerline{\includegraphics[width=1.0\textwidth]{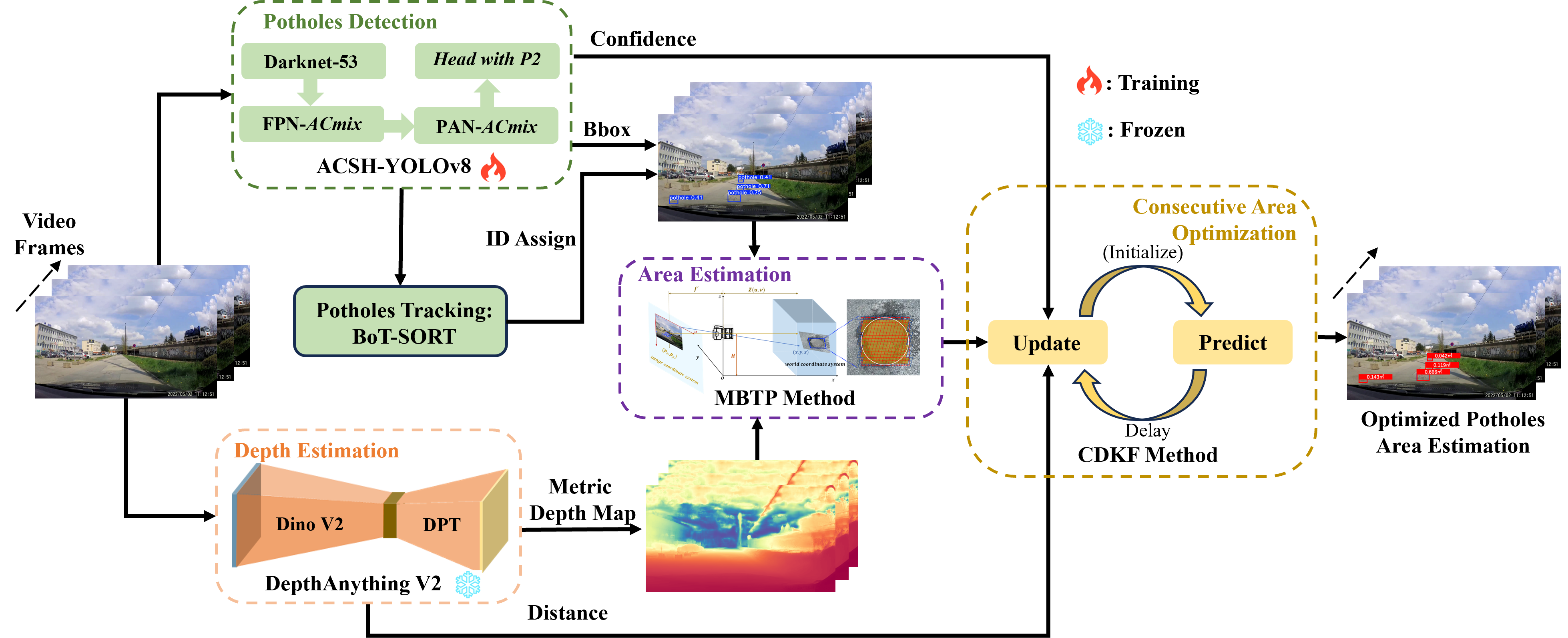}}
\caption{Overall flowchart of the proposed pothole area estimation model. The Proposed ACSH-YOLOv8 model is trainable, while DepthAnything V2 uses a pre-trained model.}
\label{structure}
\end{figure*}

The proposed framework for pothole detection and area estimation is illustrated in Fig.~\ref{structure}. First, video streams captured by the vehicle’s front camera are fed into ACSH-YOLOv8 object detection model to localize potholes and extract their bounding boxes. Next, BoT-SORT algorithm is applied to track the detected potholes across consecutive frames, assigning a unique ID to each pothole to ensure tracking consistency. Simultaneously, the video stream is processed by the pre-trained monocular metric depth estimation model, DepthAnything V2, which generates corresponding depth maps. Subsequently, by combining the object detection results with depth maps, MBTP algorithm is introduced with a pinhole camera model-based 3D mapping and faced-based method to estimate the pothole area. Finally, to enhance the robustness of the system, the potential fluctuations in the estimated area of the same pothole across consecutive frames are constrained by Kalman Filter based on Confidence Distance (CDKF), which incorporates the detection confidence and the distance between the pothole and the camera as uncertainty factors into a novel Kalman filtering algorithm, ultimately yielding an optimized pothole area estimation.
\subsection{Potholes Detection}

\begin{figure*}[htbp]
\centerline{\includegraphics[width=1.0\textwidth]{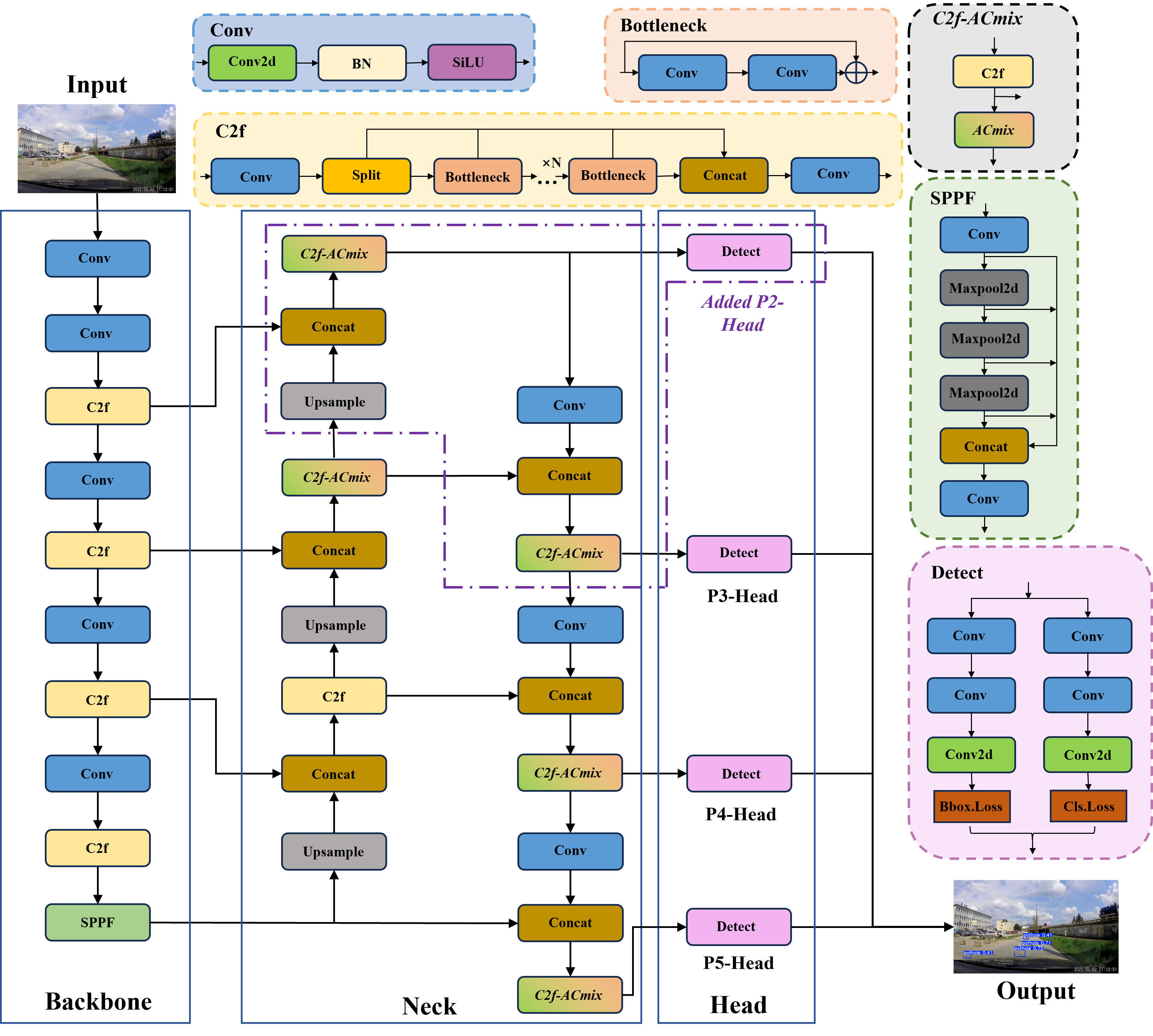}}
\caption{The ACSH-YOLOv8 model is used for pothole detection, incorporating the P2 detection head and the ACmix hybrid attention mechanism, which are highlighted in italics in the figure.}
\label{YOLO}
\end{figure*}
Object detection algorithms are commonly used for pothole detection to obtain bounding boxes of the target regions\cite{shaghouri2021real,mahalingesh2024pothole}. Given the varying sizes of potholes and their differing distances from the camera, detecting potholes accurately and consistently remains a challenging task. To address this, a novel pothole detection model named ACSH-YOLOv8 is proposed. ACSH-YOLOv8 introduces two key innovations: the addition of a small object detection head to improve detection across different scales, particularly small and distant potholes, and the integration of the ACmix attention mechanism\cite{pan2022integration} in the Neck to better focus on pothole-relevant features. The overall architecture is illustrated in Fig.~\ref{YOLO}.\par
The model consists of three main components: Backbone, Neck, and Head. The Backbone adopts the CSPDarknet structure and replaces the C3 module (used in earlier models) with the more lightweight C2f module, enhancing gradient flow while reducing computational complexity.The Neck utilizes a Feature Pyramid Network (FPN) and Path Aggregation Network (PAN) pyramid structure for multi-scale feature fusion, enabling the model to capture targets at varying scales. The Head features a decoupled design with separate branches for classification and localization, employing an anchor-free approach for bounding box prediction.\par
To further enhance detection of small-scale potholes, a dedicated P2 detection head is introduced. In contrast to standard detection heads operating on downsampled feature maps (P3: 80×80, P4: 40×40, P5: 20×20), the P2 head operates on a high-resolution 160×160 feature map. This is achieved by upsampling intermediate features in the Neck and fusing them with shallow features from the Backbone. As a result, the model retains more fine-grained visual details and significantly improves the detection of small and distant potholes, showcasing strong potential for real-world road surface analysis.\par

\begin{figure*}[htbp]
\centerline{\includegraphics[width=1.0\textwidth]{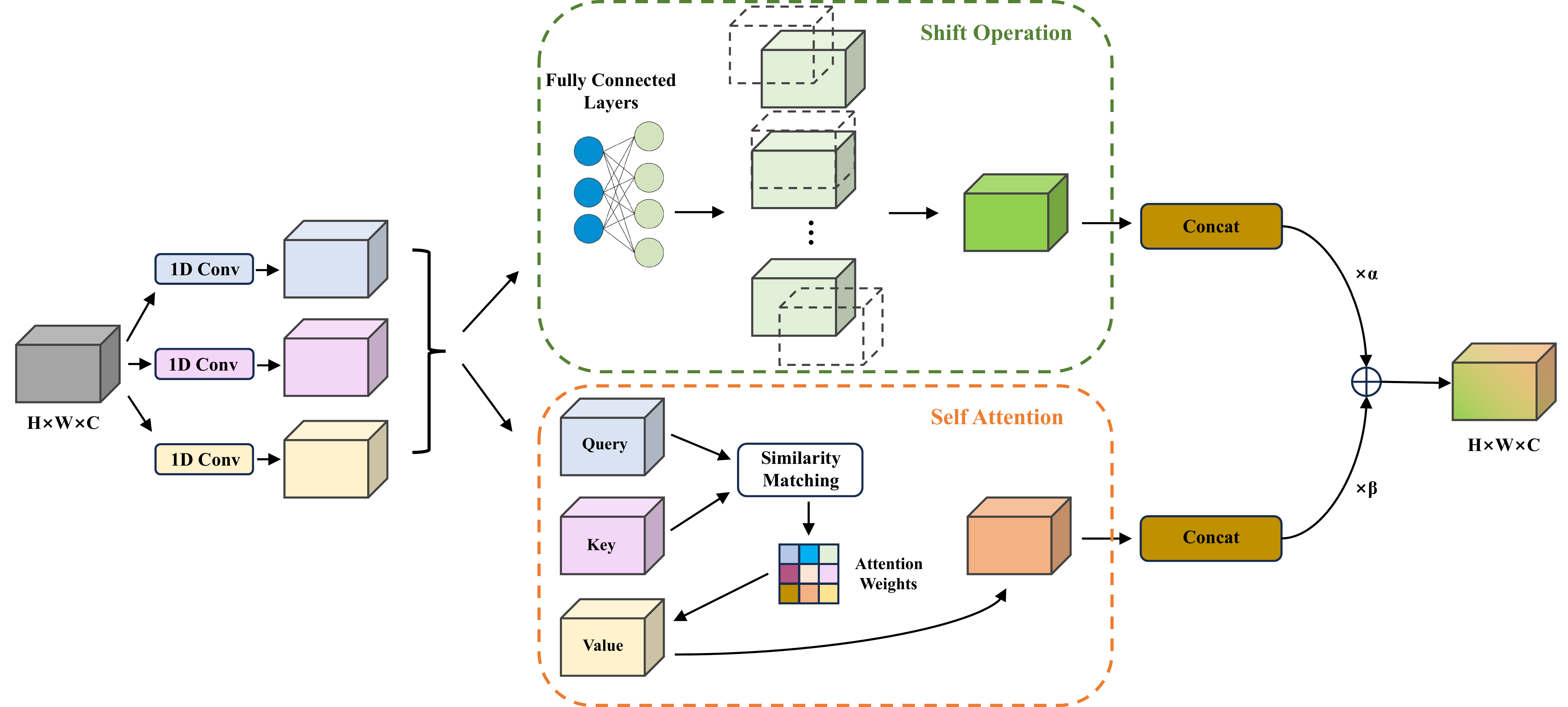}}
\caption{Schematic diagram of the ACmix hybrid attention mechanism.}
\label{ACmix}
\end{figure*}

To better focus on pothole features, we introduce the ACmix module in Neck part (including FPN and PAN), a hybrid feature extraction module that combines self-attention and convolution\cite{pan2022integration}. Its structure is shown in Fig.~\ref{ACmix}. The process begins by applying three 1×1 convolution layers to obtain three distinct feature maps. These feature maps are then processed using Shift Operation and Self Attention. The Shift Operation first uses a fully connected layer to map the features and then applies a shift operation, similar to convolution, to aggregate the features. The Self Attention mechanism divides the extracted features into Query, Key, and Value, and uses the attention mechanism to extract key information. The specific computation formulas are as follows:
\begin{equation}
\label{pairwise}
    \mathbf{F}_{att} =\! {\underset{\mathcal{N}_k(i,j)}{\rm softmax}}\!\left(\frac{\!(\mathbf{W}_q^{(l)}\!f_{ij})\!^\top\! (\mathbf{W}_k^{(l)}\!f_{ij})}{\sqrt{d}}\!\right)\! (\mathbf{W}_v^{(l)}\!f_{ij})
\end{equation}
where $\mathcal{N}_k(i,j)$ represents a local pixel region centered at pixel \((i,j)\) with a spatial width of \(k\); \(f_{ij}\) denotes the tensor input for pixel \((i,j)\), and \(\mathbf{W}_q\), \(\mathbf{W}_k\), and \(\mathbf{W}_v\) are the mapping matrices for the three corresponding features. $d$ denotes the feature dimension of $\mathbf{W}_q^{(l)}\!f_{ij}$. The final feature \(\mathbf{F}_{att}\) is obtained based on self-attention. The final output is obtained by aggregating and summing the results of the Shift Operation and Self Attention, as shown below.
\begin{equation}
\label{acmixout}
    \mathbf{F}_{out} =\alpha \mathbf{F}_{conv} + \beta \mathbf{F}_{att}
\end{equation}
where $\mathbf{F}_{conv}$ is the feature obtained from the Shift Operation, and $\alpha$ and $\beta$ are learnable weights.

\subsection{Potholes Tracking}
To track and assign consistent IDs to the same pothole across successive frames, we employ the BoT-SORT algorithm\cite{aharon2022bot}, which is well-suited for object tracking under vehicle-mounted camera motion. To ensure robust tracking, ego-motion is compensated for using sparse optical flow\cite{bouguet2001pyramidal}, which estimates global scene motion caused by camera shifts such as rolling, pitching, and translation. Specifically, the corner keypoints $\mathbf{p}_i^{k-1}$ are detected in frame $k-1$ and tracked into frame $k$ as $\mathbf{p}_i^k$ by minimizing:
\begin{equation}
\Delta \mathbf{p}_i = \arg \min_{\Delta \mathbf{p}} \sum_{\mathbf{q} \in \Omega_i^{k-1}} \left[ I_k(\mathbf{q} + \Delta \mathbf{p}) - I_{k-1}(\mathbf{q}) \right]^2
\label{flow}
\end{equation}
where $I_k$ denotes frame $k$'s intensity and $\Omega_i^{k-1}$ is a local patch around keypoint $\mathbf{p}_i^{k - 1}$. A RANSAC procedure then fits a transformation matrix $\mathbf{T}$ to the inlier flow vectors\cite{fischler1981random}, isolating the dominant camera motion. Each detected pothole bounding box $\mathbf{z}_k = (x_k, y_k, w_k, h_k)$ is compensated by transforming its center $(x_k, y_k)$ according to equation:
\begin{equation}
\begin{pmatrix}
x'_k \\
y'_k \\
1
\end{pmatrix}
=
\mathbf{T}^{-1}
\begin{pmatrix}
x_k \\
y_k \\
1
\end{pmatrix}
\label{flow_box}
\end{equation}
thereby reducing apparent motion due to vehicle movement. The updated bounding box $\mathbf{z}'_k = (x'_k, y'_k, w_k, h_k)$ then serves as input for the subsequent tracking steps.
\par
Each track is represented by an eight-dimensional Kalman filter state $\mathbf{x}_k = (x, y, w, h, \dot{x}, \dot{y}, \dot{w}, \dot{h} )^\top$, where $(x, y)$ specifies the bounding-box center, $(w, h)$ specifies the width and height, and $(\dot{x}, \dot{y}, \dot{w}, \dot{h} )$ denotes the corresponding velocities. The state is propagated using a constant-velocity transition matrix $\mathbf{F} \in \mathbb{R}^{8 \times 8}$, which can be partitioned into sub-blocks for position and velocity.\par
Given the predicted state from the previous frame, the Kalman filter uses a constant-velocity model to project the current state and its uncertainty forward. Upon receiving a new bounding box detection, the filter updates the predicted state by incorporating the measurement, adjusting both the estimate and the associated uncertainty based on the Kalman gain. This process refines the position, size, and velocity estimates of the tracked object while mitigating measurement noise and residual motion, thereby stabilizing the pothole trajectory across frames.\par
A two-stage data association procedure is then performed on the predicted states. In the first stage, detection scores exceeding a predefined threshold are labeled as high-confidence, while the rest are considered low-confidence. An assignment matrix $\mathbf{X} \in \{0,1\}^{N \times M}$ is computed by minimizing the total IoU-based cost:
\begin{equation}
\min_{\mathbf{X}} \sum_{i=1}^{N} \sum_{j=1}^{M} \left( 1 - \text{IoU}(\text{box}_i, \text{det}_j) \right)
\label{track_2stage}
\end{equation}
where $\text{IoU}$ measures the overlap between track $i$'s predicted bounding box and detection $j$. The optimization is subject to one-to-one assignment constraints, ensuring that each track is matched to at most one detection and vice versa. The Hungarian algorithm~\cite{kuhn1955hungarian} is used to solve this assignment problem. In the second stage, unmatched tracks are associated with low-confidence detections under a relaxed threshold to recover occluded or ambiguous potholes. Tracks that remain unmatched over several frames are deleted, while unmatched detections initialize new tracks. This approach enables robust identity maintenance without relying on appearance features.\par

\subsection{Monocular Depth Estimation}

\begin{figure*}[htbp]
\centerline{\includegraphics[width=1.0\textwidth]{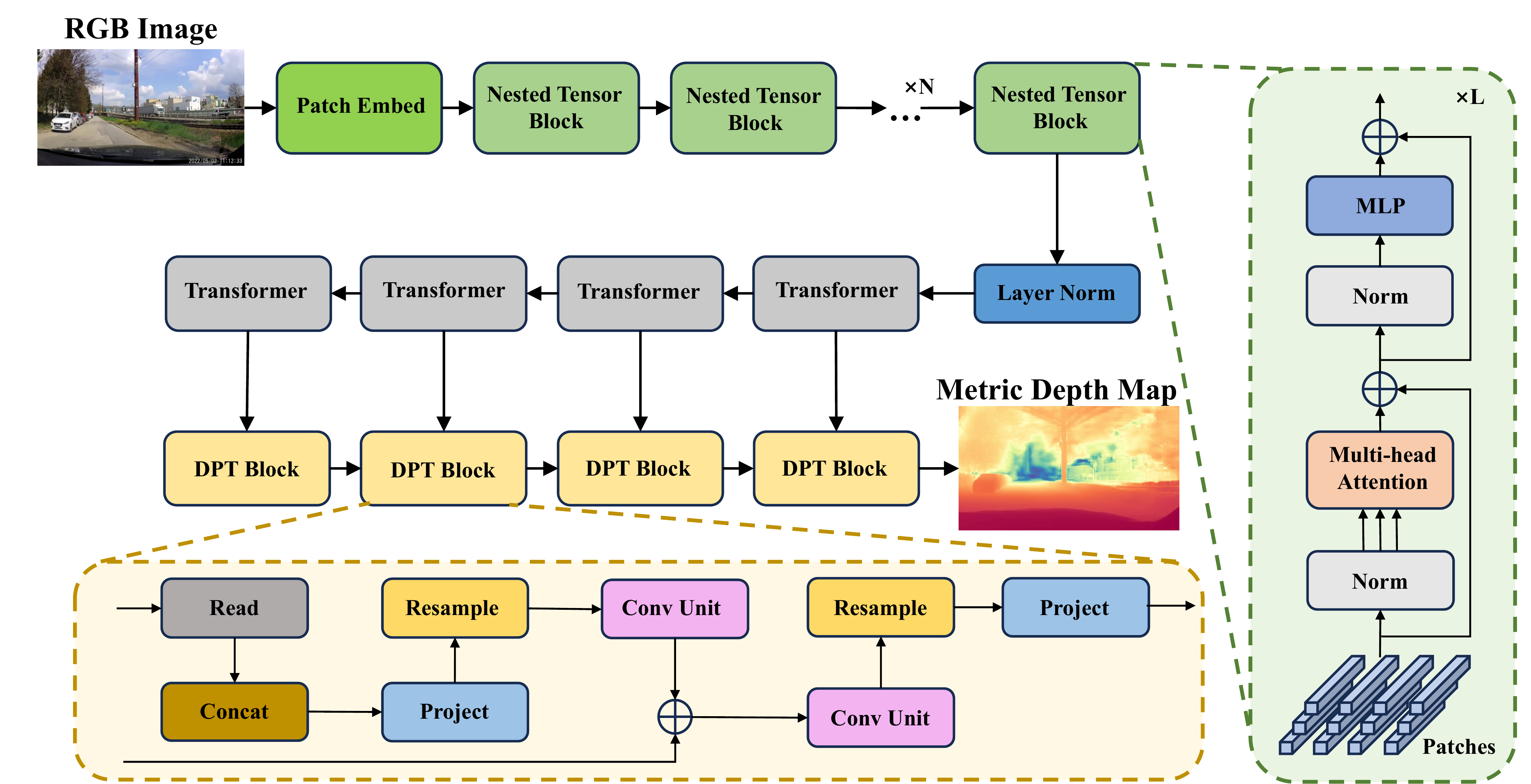}}
\caption{Schematic diagram of the monocular depth model, DepthAnything V2, used to obtain the depth values of pothole pixels.}
\label{depthmodel}
\end{figure*}

To reduce the reliance on costly 3D sensors\cite{talha2024use}, DepthAnything V2\cite{yang2024depth}, a monocular metric depth estimation model pre-trained on the KITTI dataset, is employed. It is shown to generalize well in outdoor scenes, making it suitable for estimating both the absolute distance and relative depth of potholes. The model is composed of a self-supervised visual backbone (DINO V2) and a dense prediction head (DPT). Global and local features are extracted from monocular images through transformer-based layers in the backbone, and multi-scale features are fused by the DPT head to reconstruct a full-resolution depth map. This design allows both large-scale scene structure and fine-grained pothole geometry to be captured effectively, which is essential for accurate area estimation. To enable practical deployment, a distilled version of the model is used, where the feature dimensions are reduced, and the model size is compressed from 4439.5MB to 86.2MB while maintaining high accuracy. Detailed architecture and processing flow can be found in Fig.~\ref{depthmodel}.

\subsection{Potholes Area Estimation}

\begin{figure*}[htbp]
\centerline{\includegraphics[width=1.0\textwidth]{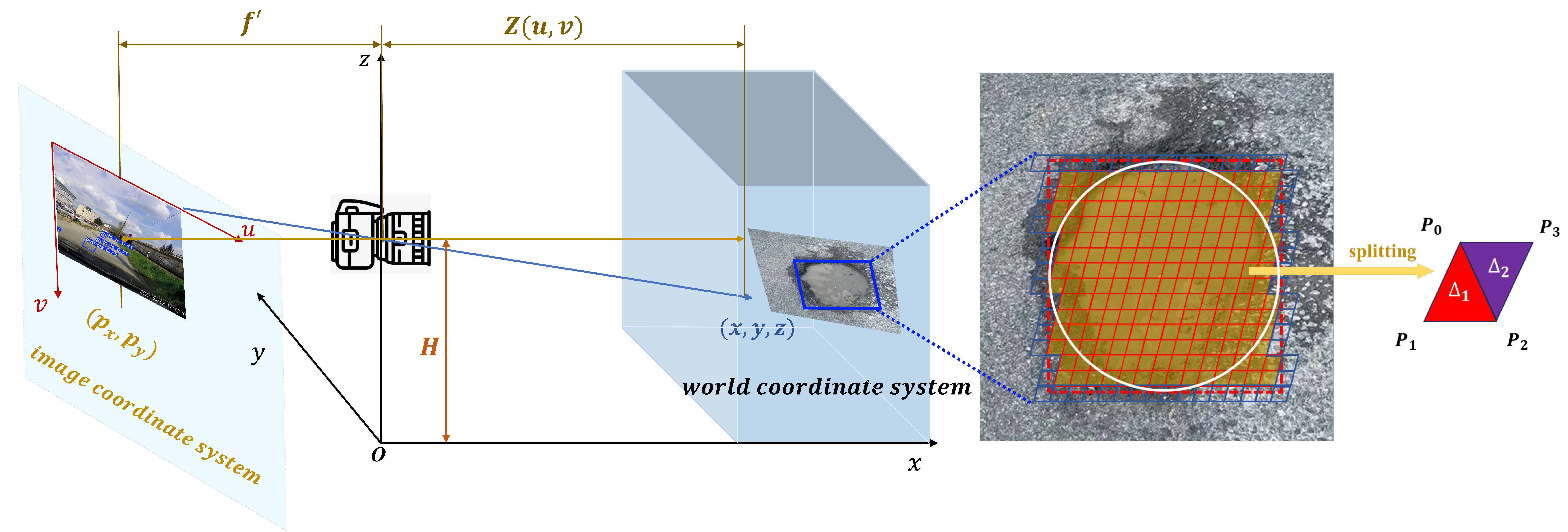}}
\caption{Schematic diagram of the proposed MBTP method.}
\label{area_calculate}
\end{figure*}

To estimate the pothole area based on the previously obtained object detection bounding boxes and monocular metric depth estimation maps, we propose the MBTP method that combines the 3D minimum bounding rectangle and the pixel-based triangular area accumulation. The process is illustrated in Fig.~\ref{area_calculate}. First, based on the pinhole camera model, the image plane pixel coordinates $(u, v)$ are converted into 3D points $(X_{u,v}, Y_{u,v}, Z_{u,v})$ in the camera coordinate system. Ignoring the effects of image distortion to simplify the model, the projection relationship from pixel $(u, v)$ to the 3D coordinates $(X_{u,v}, Y_{u,v})$ can be calculated using following formulas.
\begin{equation}
X_{u,v} = \frac{(u - p_u)}{f_u} Z_{u,v}
\label{projectionX}
\end{equation}
\begin{equation}
Y_{u,v} = \frac{(v - p_v)}{f_v} Z_{u,v}
\label{projectionY}
\end{equation}
where $p_u$ and $p_v$ represent the pixel coordinates of the camera’s optical center, while $f_u$ and $f_v$ are the focal lengths in pixels along the horizontal and vertical axes, respectively. $Z_{u,v}$ indicates the depth value corresponding to each pixel $(u, v)$ in the depth estimation map. For a given object bounding box region, denoted as $\mathcal{R}_{uv}$ in the image coordinate system, the projection of all pixels in this region onto the 3D space plane $(X, Y)$ is obtained.

\begin{equation}
\mathcal{D} = \left\{ \left( X_{u,v}, Y_{u,v} \right) \mid (u, v) \in \mathcal{R}_{uv}\right\}
\label{region_origin}
\end{equation}

To facilitate pixel-based calculations, the minimum bounding rectangle for the region is computed. By calculating the minimum and maximum values in the $\mathcal{D}$, and rounding down to avoid overestimation, the rectangular region is determined.

\begin{equation}
\mathbf{Rect}_{XY} = [{\min X},{\max X}] \times [{\min Y},{\max Y}] \quad (X,Y) \in \mathcal{D}
\label{region_rect}
\end{equation}

Within the resulting rectangular region $\mathbf{Rect}_{XY}$, the area is subdivided into segments formed by adjacent groups of 2×2 pixels. The 3D projection coordinates of these four neighboring points are represented as $
P_0 = \bigl(X_{u,v},\, Y_{u,v}\bigr)$, $
P_1 = \bigl(X_{u+1,v},\, Y_{u+1,v}\bigr)$, $P_2 = \bigl(X_{u,v+1},\, Y_{u,v+1}\bigr)$, $P_3 = \bigl(X_{u+1,v+1},\, Y_{u+1,v+1}\bigr)$. To simplify further area calculations, each diamond is split into two triangles, $\Delta_1(P_0, P_1, P_2)$ and $\Delta_2(P_0, P_2, P_3)$. The area of each triangle is then calculated using the vector cross product formula as described below.
\begin{equation}
\text{Area}(\Delta) 
= \frac{1}{2} \bigl | (x_2 - x_1)(y_3 - y_1) - (y_2 - y_1)(x_3 - x_1) \bigr |
\label{area_tri}
\end{equation}
In this context, $x$ and $y$ denote the coordinate components of the three vertices of each triangle. By summing the areas of the two triangles, we obtain the area of the corresponding 2×2 pixel block, denoted as $A^{\text{patch}}_{unv}$. The total area of the rectangular region $S_{\text{final}}$ is then determined by accumulating the areas of all valid blocks within that region. Since most potholes are elliptical, the final area is approximated by multiplying the rectangular area by a coefficient of $\frac{\pi}{4}$.
\begin{equation}
A^{\text{patch}}_{u,v} = \text{Area}(\Delta_1) + \text{Area}(\Delta_2)
\label{area_rect}
\end{equation}
\begin{equation}
S_{\text{final}} 
= \left( 
    \sum_{\substack{(u,v)\in\mathcal{R}_{uv} \\ P_i \in \text{Rect}_{XY}}} 
    A^{\text{patch}}_{u,v} 
\right) 
\times \frac{\pi}{4}
\label{area_sum}
\end{equation}

\subsection{Consecutive Frame Area Optimization}

To address the uncertainty and noise interference in estimating pothole areas from video frames, this paper proposes Kalman Filter based on Confidence and Distance (CDKF), a robust estimation method based on Kalman filtering. The method leverages the Kalman filter's predict-update mechanism to recursively estimate and dynamically smooth the pothole area state. Additionally, it incorporates an adaptive measurement noise adjustment strategy based on detection confidence and distance to achieve more robust estimation in clear and dark environments.\par
For an individual pothole, it is assumed that its area remains constant across consecutive frames, conforming to a constant state model. Due to unavoidable noise introduced during processes such as bounding box detection and monocular depth estimation, process noise covariance is incorporated to update the state uncertainty. In the prediction step, the state and covariance are updated as follows:
\begin{equation}
\mathbf{A}_{k|k-1} = \mathbf{A}_{k-1}, \mathbf{P}_{k|k-1} = \mathbf{P}_{k-1|k-1} + \mathbf{Q}
\label{area_kal_update}
\end{equation}
where $\mathbf{A}_k$ denotes the estimated pothole area in the $k$th frame, $\mathbf{P}$ represents the state uncertainty, and $\mathbf{Q}$ is the process noise covariance. During the measurement update phase, the predicted state is adjusted using the pothole area measurement obtained from the current frame detection. The update of the Kalman filter result relies on the Kalman gain, which is computed using the following formula:
\begin{equation}
\mathbf{K}_{k-1} = \frac {\mathbf{P}_{k|k-1}}{\mathbf{P}_{k|k-1} + \mathbf{R}_{k-1}}
\label{area_kal_gain}
\end{equation}
In this context, $\mathbf{K}$ represents the Kalman gain, which determines the degree of trust placed in the current observation, while $\mathbf{R}$ denotes the measurement noise covariance. Compared to conventional Kalman filter methods, a key innovation of this approach is the dynamic adjustment of $\mathbf{R}$. This adjustment considers two crucial factors affecting measurement accuracy: the confidence level of the bounding box from the object detection algorithm and the distance between the pothole and the camera. Specifically, a higher bounding box confidence and a closer proximity to the camera both contribute to higher measurement accuracy. Based on these considerations, the measurement noise covariance is determined using the following equation:
\begin{equation}
\mathbf{R}_{k-1} = \frac {\lambda }{c} + \theta \cdot \max\{d,d_0\}
\label{area_kal_R}
\end{equation}
where $d$ denotes the distance from the center of the pothole bounding box to the camera, and $c$ represents the confidence level of the detection bounding box. Since the pothole area estimates are generally more reliable at closer ranges, a trusted distance range $d_0$ is defined, within which $\mathbf{R}$ remains unaffected by changes in distance. The parameters $\lambda$ and $\theta$ serves as tuning factors to balance the influences of both the confidence and distance on $\mathbf{R}$. Ultimately, the Kalman filter state update and covariance correction are expressed as follows:

\begin{equation}
\mathbf{A}_{k} = \mathbf{A}_{k|k-1} + \mathbf{K}_{k - 1}(z_k - \mathbf{A}_{k|k-1})
\label{area_kal_A}
\end{equation}

\begin{equation}
\mathbf{P}_{k} = (1 - \mathbf{K}_{k - 1})\mathbf{P}_{k|k-1}
\label{area_kal_P}
\end{equation}
In the above equation, $(z_k - \mathbf{A}_{k|k-1})$ represents the measurement residual (innovation), whose magnitude reflects the inconsistency between the observed value and the prior prediction.\par
\section{Experimental Evaluations}
\subsection{Dataset Construction}
\subsubsection{Dataset Configuration}
The dataset used in this study originates from previous research by Bu{\v{c}}ko\cite{buvcko2022computer}. It consists of images captured by a camera mounted on the front side of a vehicle. The overall dataset collection includes several datasets captured under different times of day and weather conditions. For this study, two representative datasets were selected: the Clear Road Dataset, representing clear weather conditions, and the Dark Road Dataset, representing low-light conditions such as dusk and nighttime. These two datasets were used separately for training and evaluation.\par
The Clear Road Dataset contains 1,052 images, including 1,896 pothole instances and 232 manhole instances. The Dark Road Dataset comprises images taken during dusk and nighttime, with 250 and 310 images respectively, totaling 560 images. This subset includes 506 pothole instances and 95 manhole instances. All images have a resolution of 1920×1080.\par

\begin{figure}[htbp]
\centerline{\includegraphics[width=0.5\textwidth]{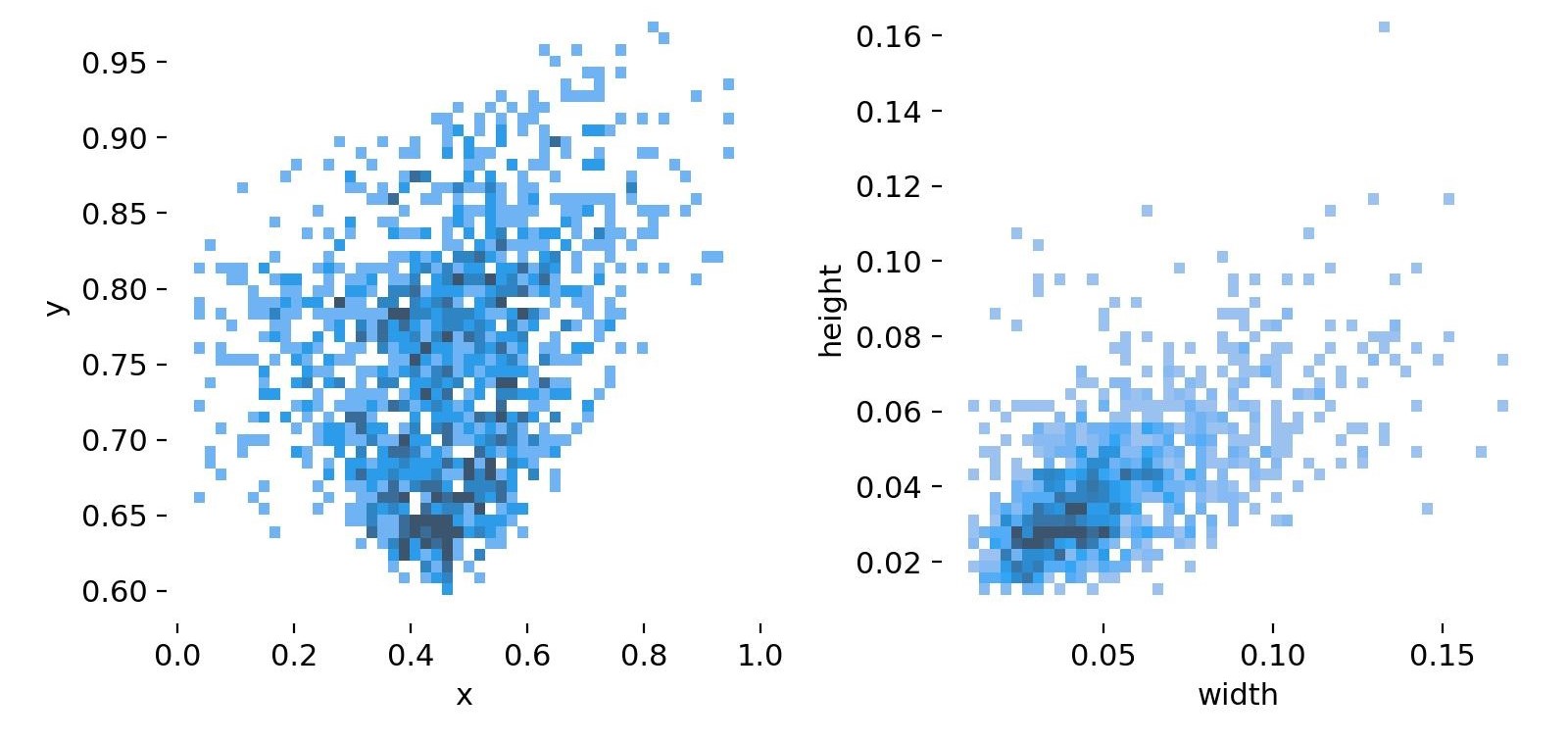}}
\caption{Distribution of object center and size in Clear Road Dataset.}
\label{pothole_size}
\end{figure}

\begin{figure}[htbp]
\centerline{\includegraphics[width=0.5\textwidth]{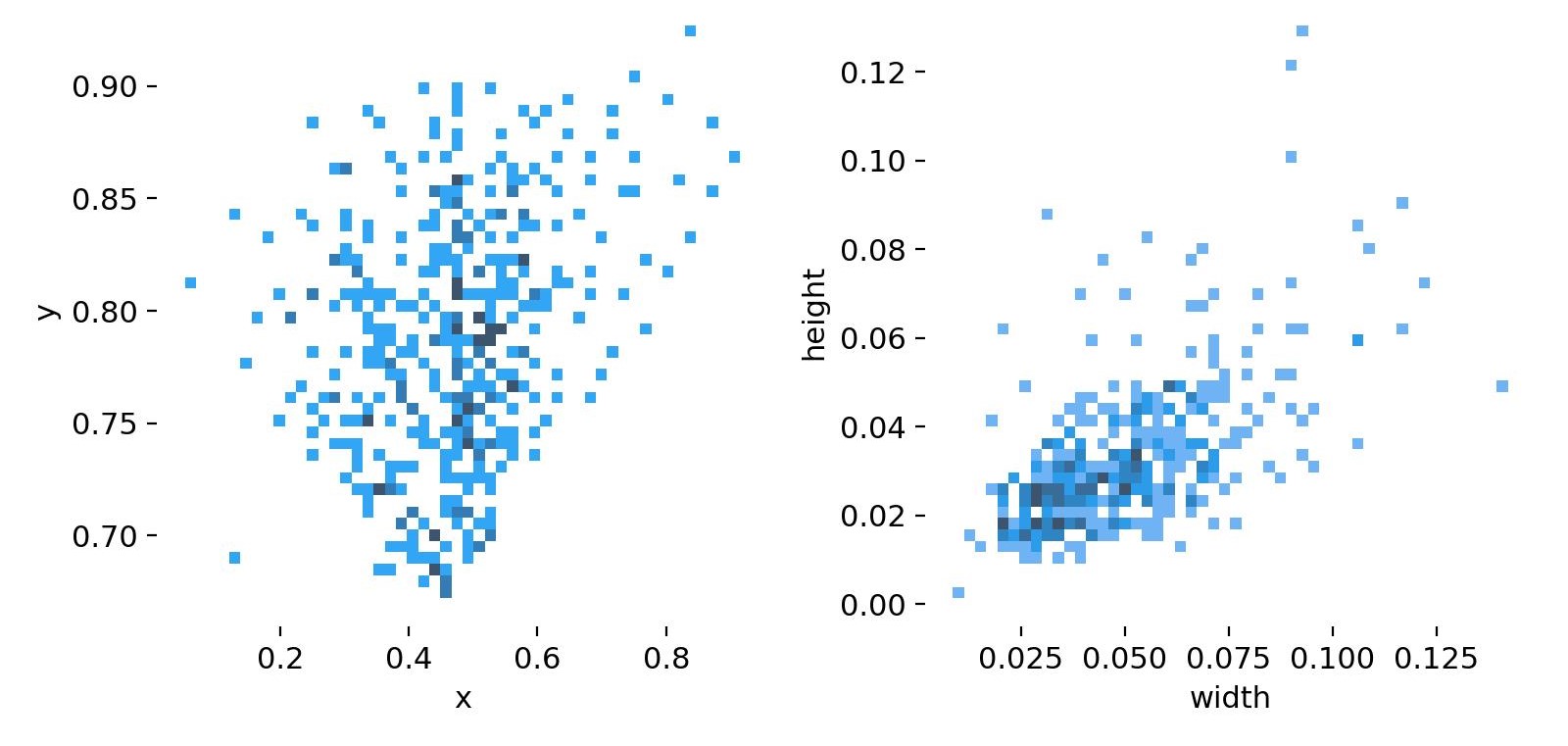}}
\caption{Distribution of object center and size in Dark Road Dataset.}
\label{pothole_location}
\end{figure}

The statistical distributions of pothole targets in the Clear Road Dataset and Dark Road Dataset are shown in Fig.~\ref{pothole_size} and Fig.~\ref{pothole_location}, respectively. These figures provide a representative overview of the potholes captured under different conditions in reality. In each figure, the left side illustrates the distribution of pothole center points across the image, while the right side shows the normalized width and height distributions of the detected targets. The two datasets exhibit similar patterns. Most potholes are relatively small in size, with their widths and heights concentrated between 0.02 and 0.06 along the normalized horizontal and vertical axes. This highlights the inherent challenge of small object detection in pothole scenarios. Due to occlusion from the vehicle’s front end, the lower portion of the image becomes a detection blind spot, resulting in most potholes being detected in the upper region. At closer distances, potholes are primarily detected near the center of the image, as detection on the sides is often hindered by occlusions or poor lighting conditions. The vertical distribution of potholes ranges from 0.6 to 0.95, indicating that the dataset captures potholes at varying distances—from far to near. This makes it suitable for validating the proposed continuous-frame pothole area estimation and optimization algorithm.\par

\subsubsection{Data Augmentation}
In order to improve the robustness and accuracy of the pothole detection model, we apply extensive data augmentation techniques as follows. Firstly, data augmentation methods include horizontal flipping, applied to half of the images, leveraging the inherent symmetry of potholes to effectively double the dataset. To simulate variations in lighting and environmental conditions, images are transformed into the HSV color space with controlled adjustments: hue varies by 1.5\%, saturation by 70\%, and brightness by 40\%. Additionally, image scaling within a 50\% range is implemented to mimic potholes appearing at different distances and sizes, enhancing the model's flexibility. Furthermore, scaling at various distances imitates potholes captured from diverse viewpoints, improving the model’s responsiveness to varying object sizes. The Mosaic augmentation technique proves particularly beneficial for improving the detection accuracy of smaller potholes. It combines four randomly selected images into one composite image, enriching the context and diversity presented to the model during training. These strategies collectively aim to boost the generalizability and reliability of the pothole detection model across varying operational scenarios.\par

\subsection{Training and Evaluation of Pothole Detection}
\subsubsection{Pothole Detection Evaluation Metrics}
To evaluate the performance of pothole detection and demonstrate the effectiveness of proposed ACSH-YOLOv8 over the baseline model, we introduce multiple evaluation metrics, including precision, recall, AP (50), AP (50-95), and GFLOPs. Since the primary focus is on pothole detection, the detection results for manholes in the dataset are not included in the evaluation. The formulas for calculating precision and recall are as follows:
\begin{equation}
P = \frac {TP}{TP + FP}, R = \frac {TP}{TP + FN}
\label{PR}
\end{equation}
where $P$ represents precision, $R$ represents recall, $TP$ represents the number of correctly detected samples, $FP$ denotes the number of falsely detected samples, and $FN$ refers to the number of missed detections. Setting the Intersection over Union (IoU) threshold at 70\%, a predicted bounding box is considered a true positive only if its IoU with the ground truth exceeds this threshold. To balance precision and recall, we use the F1-score and AP (Average Precision) metric, which are calculated as follows:

\begin{equation}
F_1 = \frac{2 \cdot P \cdot R}{P + R}
\label{F1}
\end{equation}

\begin{equation}
AP = \int_{0}^{1} P(R) \, \mathrm{d}R
\label{AP}
\end{equation}

The F1 score is a harmonic mean of precision and recall, providing a balanced measure for evaluating classification models, especially in imbalanced datasets. The AP metric represents the integral area under the Precision-Recall (P-R) curve. AP is calculated based on different IoU thresholds, with two commonly used variants: AP(50) and AP(50-95). AP(50) refers to the AP score computed with a fixed IoU threshold of 0.5, providing a more lenient evaluation of detection performance. In contrast, AP(50-95) is a more stringent metric that averages AP scores calculated at IoU thresholds ranging from 0.5 to 0.95, in increments of 0.05. This comprehensive evaluation better reflects the model’s localization accuracy. Additionally, to assess the computational complexity of the model, we use GFLOPs, which measures the number of floating-point operations required for a single forward inference.\par

\subsubsection{Model Training Configuration}

During model training, the input image size is set to 1080×1080 to ensure that fine details of potholes can be effectively captured. The batch size is set to 4. Training is performed using the SGD momentum optimizer, with a momentum value of 0.937 and a weight decay of 0.0005. The initial learning rate is set to 0.01, gradually decreasing to a final learning rate of 0.0001. A warmup training strategy is applied, with the first 3 epochs dedicated to warmup training. During this phase, the momentum is set to 0.8, and the bias learning rate is 0.1.\par
Both training and validation are conducted using the PyTorch 2.3.1 deep learning framework with CUDA version 12.6. The hardware setup includes an RTX 4090 GPU with 24GB of VRAM and 16 vCPU Intel Xeon Gold 6430 processor.\par

\subsection{Evaluation of Pothole Area Estimation and Optimization}
\subsubsection{Pothole Area Estimation Evaluation Metrics}
For an ideal pothole area estimation algorithm, the estimates for the same pothole should be similar across different frames, demonstrating that the model can reliably predict potholes regardless of their position and size. Given the absence of ground truth measurements, we evaluate the estimation method’s accuracy and consistency using statistical measures based on multiple observations of the same pothole. To assess this consistency, we introduce three evaluation metrics: Mean Absolute Error (MAE), Coefficient of Variation (CV), and adjacent frame differences (AFD). In all cases, lower values indicate better performance. Mean Absolute Error (MAE) measures the average deviation between each estimated area and the mean of all estimates for a single pothole.
\begin{equation}
\text{MAE} = \frac{1}{N}\sum_{k=1}^{N}\left|A_k - \bar{A}\right|
\label{MAE}
\end{equation}
In the above, $\bar{A}$ represents the mean area estimate for a specific pothole, calculated as follows.

\begin{equation}
\bar{A} = \frac{1}{N}\sum_{k=1}^{N}A_k
\label{abar}
\end{equation}

Next, to assess the relative dispersion and consistency of the area estimates, we introduce the Coefficient of Variation (CV) metric.

\begin{equation}
\text{CV} = \frac{\sqrt{\frac{1}{N}\sum_{k=1}^{N}(A_k-\bar{A})^2}}{\bar{A}}
\label{cv}
\end{equation}

Additionally, to quantify the variation between consecutive frame estimates and reflect the smoothness of the filter output, we incorporate a metric based on adjacent frame differences (AFD).

\begin{equation}
\text{AFD} = \frac{1}{N-1}\sum_{k=2}^{N}|A_k - A_{k-1}|
\label{AFD}
\end{equation}

To further assess the reliability of our improved Kalman filtering method, we introduce the Normalized Innovation Squared (NIS) metric to evaluate the filter's internal consistency, reflecting how well the noise model matches the actual measurement data. The NIS is calculated using a specific formula as follows.

\begin{equation}
\text{NIS} = \nu_k^\top \left(\mathbf{P}_{k|k-1} + \mathbf{R}_k\right)^{-1}\nu_k
\label{nis}
\end{equation}
where $\nu_k$ represents the difference between the observed measurement and the predicted value at each time step. $\mathbf{P}_{k|k-1}$ represents the uncertainty level of the predicted value $\mathbf{A}_{k|k-1}$ and reflects the filter’s uncertainty estimate during the prediction step, while $\mathbf{R}_k$ indicates the filter’s expected level of uncertainty in the measurement data. Theoretically, if the filter correctly models both the system and measurement noise, the NIS should statistically follow a chi-square distribution with an expected value related to the measurement dimension. In this study, since the area measurement is one-dimensional, the NIS should ideally be as close to 1 as possible.
\subsubsection{Kalman Filter Bayesian Parameter Optimization}
In optimizing the area estimation results using Kalman filtering, the calculation of noise covariance is crucial. In our study, the noise covariance consists of two components: confidence and pothole distance, as defined in Eq.~\ref{area_kal_R}. To determine the optimal weights lambda and theta for these two noise factors, we employ a Bayesian optimization algorithm to maximize overall filtering performance.
\begin{equation}
    J(\lambda, \theta) = 10 \cdot \text{MAE} + \text{CV} + \text{AFD} + \text{NIS}
\label{J_define}
\end{equation}
To comprehensively evaluate the filter’s performance, we define a combined evaluation metric $J$, which integrates multiple indicators. The calculation methods for MAE, CV, AFD, and NIS are described in the previous section. Since MAE has a relatively small magnitude compared to the other metrics, we multiply it by a factor of 10 for better balance. These indicators depend on the filtering process, which in turn is influenced by $\mathbf{R}(\lambda, \theta)$, making $J$ an implicit function of $\lambda$ and $\theta$. The objective is to find the optimal parameters that minimize $J$.
\begin{equation}
    (\lambda^*, \theta^*) = \arg\min J(\lambda, \theta)
\label{J_optimaze}
\end{equation}

Specifically, we use the BayesianOptimization method from Python’s bayes$\_$opt library. The search range for both parameters is set between 0 and 2, with an initial exploration of five trials followed by 30 iterations to determine the optimal values.

\subsection{Potholes Detection Results}

To further evaluate the effectiveness of our pothole detection model, we conduct a series of comparative experiments under Clear Road Dataset and Dark Road Dataset, as shown in Table.~\ref{detect_table}. To ensure objective comparison, this study selects multiple representative baseline models for evaluation. For the benchmark YOLOv8 series, we employ the baseline YOLOv8n\cite{kumari2023yolov8}, its lightweight variant YOLOv8n-ghost\cite{hussein2024real}, and the YOLOv8s\cite{kumari2023yolov8} model to conduct comprehensive comparisons. The models were assessed using precision, recall, F1-score, AP(50), AP(50-95), and GFLOPs.\par

\begin{table*}
\centering
\renewcommand{\arraystretch}{1.5}  
\caption{Comparison of different models results for pothole detection.}
\begin{threeparttable}
\resizebox{\linewidth}{!}{
\begin{tabular}{c|c|cccccc}
\toprule
\textbf{Dataset} & \textbf{Pothole Detection Model} & \textbf{Precision $\uparrow$} & \textbf{Recall $\uparrow$} & \textbf{F1-score $\uparrow$} & \textbf{AP (50) $\uparrow$} & \textbf{AP (50-95) $\uparrow$} & \textbf{GFLOPs $\downarrow$} \\
\midrule
\multirow{8}{*}{\centering Clear Road} 
& YOLOv3-tiny\cite{adarsh2020yolo} & \textbf{72.5\%} & 58.7\% & 64.9\% & 69.9\% & 26.7\% & 18.9 \\
& YOLOv3-spp\cite{liu2024improved} & 63.5\% & 60.6\% & 62.0\% & 64.9\% & 25.6\% & 12.0 \\
& YOLOv5n\cite{ajmera2022real} & 71.0\% & 64.8\% & 67.8\% & 69.0\% & 26.9\% & 7.1 \\
& YOLOv6n\cite{li2022yolov6} & 53.7\% & 68.1\% & 60.1\% & 62.1\% & 25.7\% & 11.9 \\
& YOLOv8n\cite{kumari2023yolov8} & 55.0\% & 71.3\% & 62.1\% & 69.0\% & 28.5\% & 8.2 \\
& YOLOv8n-ghost\cite{hussein2024real} & 66.8\% & 64.8\% & 65.8\% & 65.5\% & 24.0\% & \textbf{5.0} \\
& YOLOv8s\cite{kumari2023yolov8} & 71.2\% & 64.4\% & 67.6\% & 72.1\% & 28.6\% & 28.4 \\
& \textbf{ACSH-YOLOv8n} & 67.9\% & \textbf{71.8\%} & \textbf{69.8\%} & \textbf{76.6\%} & \textbf{28.7\%} & 13.4 \\
\midrule
\multirow{8}{*}{\centering Dark Road} 
& YOLOv3-tiny\cite{adarsh2020yolo} & 69.8\% & 55.4\% & 61.8\% & 59.0\% & 20.3\% & 18.9 \\
& YOLOv3-spp\cite{liu2024improved} & \textbf{81.1\%} & 57.1\% & 67.0\% & 67.3\% & 26.4\% & 12.0 \\
& YOLOv5n\cite{ajmera2022real} & 65.8\% & 61.8\% & 63.7\% & 62.8\% & 24.6\% & 7.1 \\
& YOLOv6n\cite{li2022yolov6} & 63.1\% & 44.6\% & 52.3\% & 48.1\% & 15.1\% & 11.9 \\
& YOLOv8n\cite{kumari2023yolov8} & 71.9\% & 59.5\% & 65.1\% & 64.4\% & 25.8\% & 8.2 \\
& YOLOv8n-ghost\cite{hussein2024real} & 60.2\% & 60.7\% & 60.4\% & 66.3\% & 24.7\% & \textbf{5.0} \\
& YOLOv8s\cite{kumari2023yolov8} & 53.6\% & \textbf{67.9\%} & 59.9\% & 68.5\% & \textbf{27.0\%} & 28.4 \\
& \textbf{ACSH-YOLOv8n} & 73.3\% & 62.5\% & \textbf{67.5\%} & \textbf{72.2\%} & \textbf{27.0\%} & 13.4 \\
\bottomrule
\end{tabular}
}
\begin{tablenotes}
		\item Note: The best results values for each metric are highlighted in bold.
\end{tablenotes}
\end{threeparttable}
\label{detect_table}
\end{table*}

\begin{figure*}[htbp]
\centerline{\includegraphics[width=1.0\textwidth]{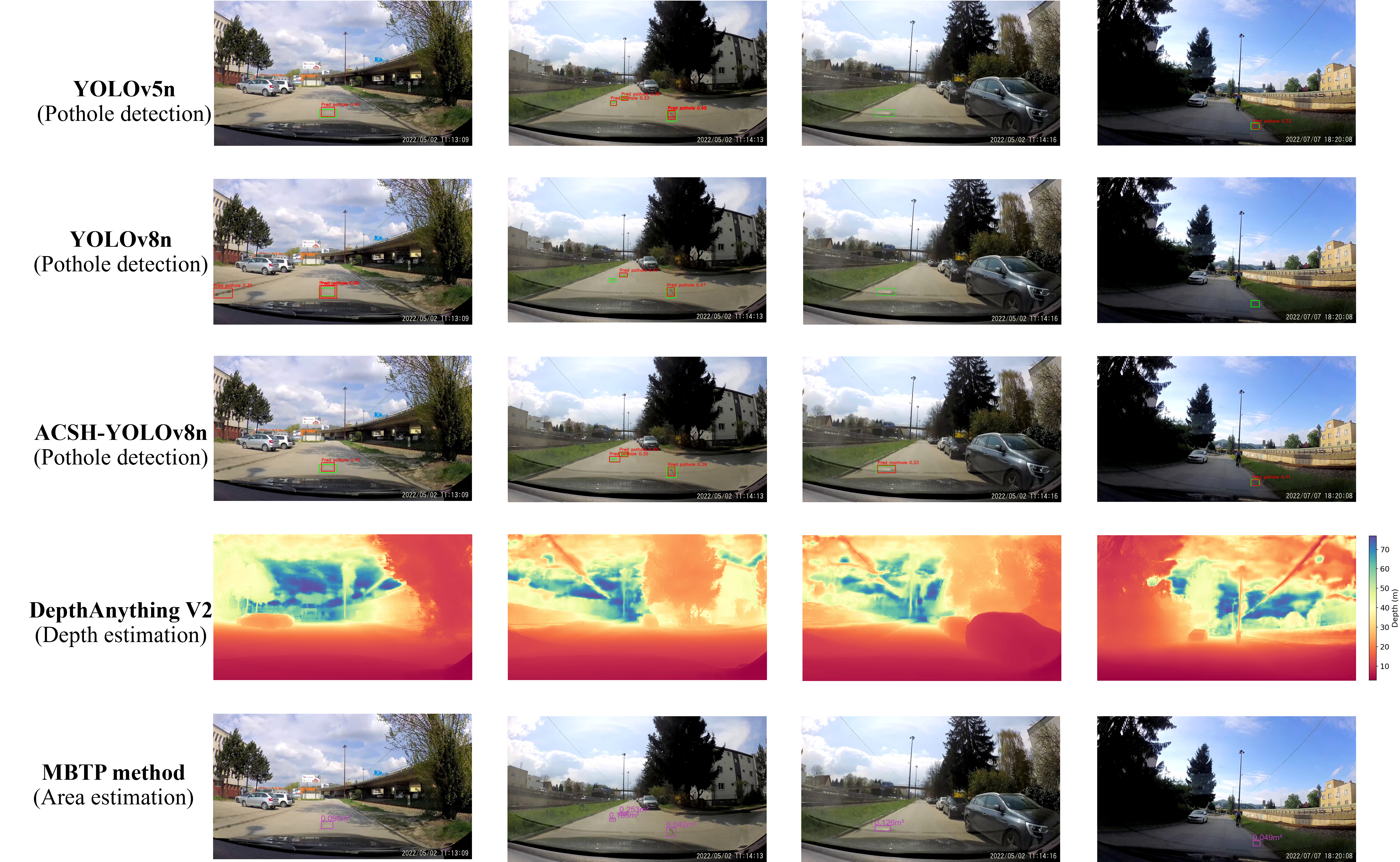}}
\caption{Comparison of pothole area estimation workflows under Clear Road Dataset. The first, second and third rows show pothole detection results using YOLOv5n, YOLOv8n and the proposed ACSH-YOLOv8n models, respectively, where red boxes indicate predicted bounding boxes and green boxes represent ground truth. The fourth row displays the monocular metric depth estimation results generated by the DepthAnything V2 model. The fifth row presents the estimated pothole areas obtained by combining detection and depth information using the proposed MBTP method.}
\label{all_show_clear}
\end{figure*}

\begin{figure*}[htbp]
\centerline{\includegraphics[width=1.0\textwidth]{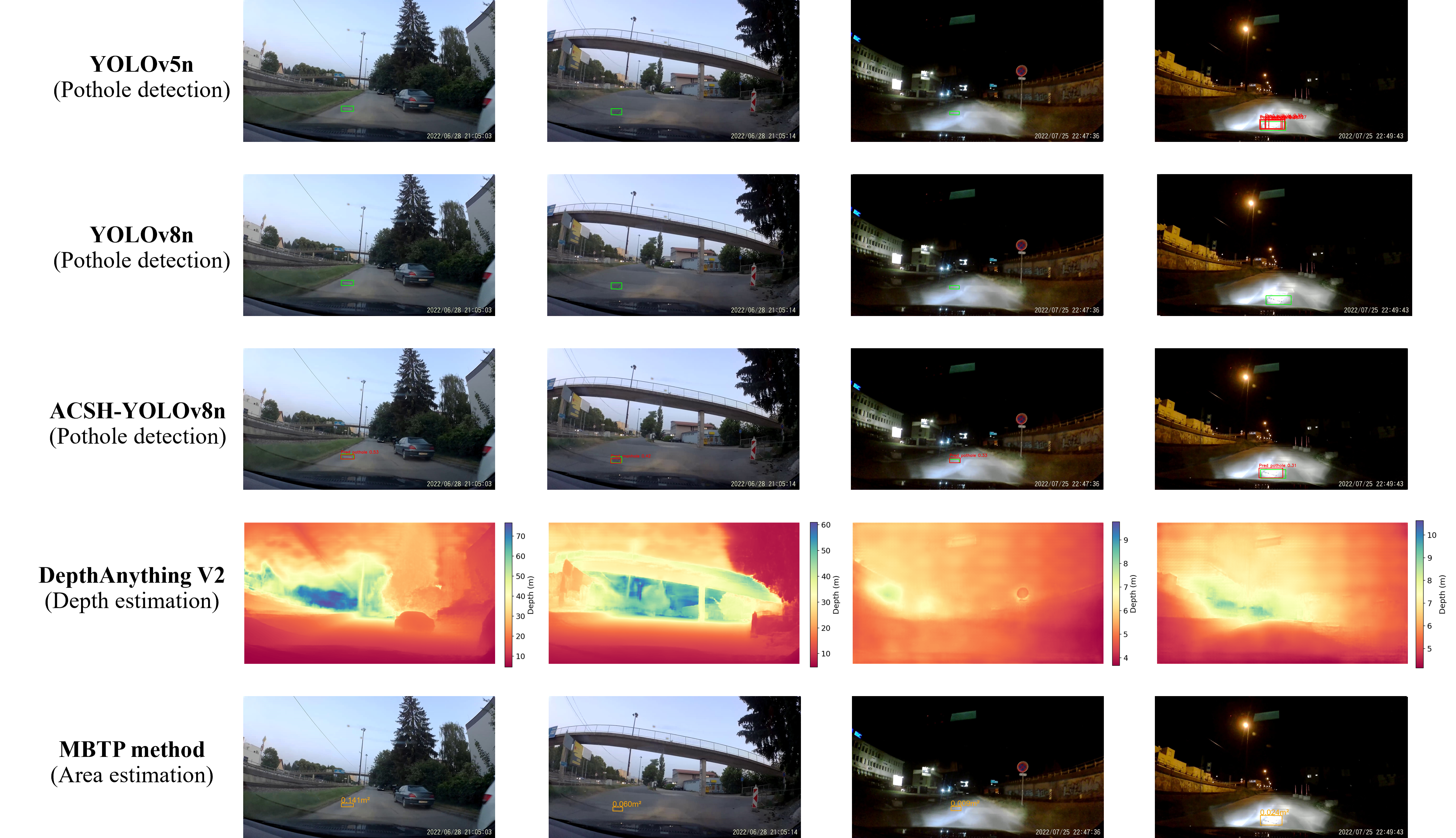}}
\caption{Comparison of pothole area estimation workflows under Dark Road Dataset. The first, second and third rows show pothole detection results using YOLOv5n, YOLOv8n and the proposed ACSH-YOLOv8n models, respectively, where red boxes indicate predicted bounding boxes and green boxes represent ground truth. The fourth row displays the monocular metric depth estimation results generated by the DepthAnything V2 model. Due to varying depth visibility across images under low-light conditions, the measurement range differs. Therefore, a scale bar is provided on the right side of each depth map for reference. The fifth row presents the estimated pothole areas obtained by combining detection and depth information using the proposed MBTP method.}
\label{all_show_dark}
\end{figure*}

The results show that on the Clear Road Dataset, the proposed ACSH-YOLOv8n model achieves the best performance in recall, F1-score, AP(50), and AP(50-95). It reaches an F1-score of 69.8\% and an AP(50) of 76.6\%, outperforming the second-best YOLOv5n by 2\% in F1-score and YOLOv8s by 4.5\% in AP(50). On the Dark Road Dataset, ACSH-YOLOv8n also achieves the highest scores in key metrics, including F1-score, AP(50), and AP(50-95). It records an AP(50) of 72.2\%, surpassing the next best model, YOLOv8s, by 3.7\%. Overall, while YOLOv3-tiny and YOLOv3-spp attain the highest precision on the two datasets respectively, their recall is significantly lower, resulting in weaker overall performance. The YOLOv8s model delivers competitive AP(50-95) scores comparable to ACSH-YOLOv8n across both datasets, along with strong performance in other metrics, reflecting the robustness of the YOLOv8 architecture. However, YOLOv8s has more than twice the number of parameters compared to ACSH-YOLOv8n. This further demonstrates that the proposed model not only achieves high detection performance but also maintains a lightweight structure suitable for deployment. The performance gains are primarily attributed to architectural improvements rather than merely increasing model size.\par

The visual comparisons of detection performance on the Clear Road Dataset and Dark Road Dataset are shown in the first three rows of Fig.~\ref{all_show_clear} and Fig.~\ref{all_show_dark}, respectively. In Fig.~\ref{all_show_clear}, the YOLOv5n model exhibits issues such as duplicate detections in the second image, where multiple bounding boxes are assigned to the same pothole, and missed detections in the third image. The YOLOv8n model produces a false positive in the first image by misidentifying a road step as a pothole. Additionally, in the second, third, and fourth images, it fails to detect several potholes, particularly small ones. In contrast, ACSH-YOLOv8n model demonstrates significantly improved performance, effectively detecting small potholes and achieving a higher recall rate. It also shows better alignment with the actual pothole contours, indicating strong potential for enhancing safety in autonomous driving scenarios. In Fig.~\ref{all_show_dark}, both YOLOv5n and YOLOv8n miss detections in the first three images. In the fourth image, YOLOv5n detects the same pothole three times with separate bounding boxes, while YOLOv8n continues to miss the pothole entirely. In comparison, ACSH-YOLOv8n model successfully addresses all these issues, demonstrating robust performance even in low-light conditions such as dusk or nighttime. This highlights the model’s strong adaptability and potential for improving pothole detection reliability in challenging lighting environments, ultimately contributing to safer driving.\par


\subsection{Area Estimation and Optimization Results}
As the only available related approach in prior studies\cite{wang2024novel}, the Corner Point (CP) method is used as a baseline to validate the effectiveness of our proposed MBTP method in area estimation. The evaluation is conducted using three key metrics: MAE, CV, and AFD, with the results presented in the upper section of Table.~\ref{area_table}. Our method consistently outperforms the corner point method across all three metrics, demonstrating its robustness and reliability.\par

\begin{table}
\centering
\renewcommand{\arraystretch}{1.5}  
\caption{Comparison of different area estimation algorithms and consecutive frame optimization strategies on the two datasets.}\label{table1}
\begin{threeparttable}
\resizebox{\linewidth}{!}{
\begin{tabular}{c|cccc|cccc}
\toprule
\textbf{Dataset} & \textbf{MBTP}  &  \textbf{KF} & \textbf{Confidence} & \textbf{Distance} &\textbf{MAE $\downarrow$} & \textbf{CV $\downarrow$} & \textbf{AFD $\downarrow$} & \textbf{NIS $\downarrow 1$} \\
\midrule
\ & & & & & 0.168 & 0.379 & 0.143 & / \\
\ & \CheckmarkBold & & & & 0.147 & 0.262 & 0.123 & / \\
Clear Road & \CheckmarkBold & \CheckmarkBold & \CheckmarkBold & &0.054 & 0.199 & 0.035 & \textbf{1.404} \\
\ & \CheckmarkBold & \CheckmarkBold &  & \CheckmarkBold &\textbf{0.036} & \textbf{0.111} & 0.056 & 2.120 \\
\ & \CheckmarkBold & \CheckmarkBold & \CheckmarkBold & \CheckmarkBold & \textbf{0.038} & \textbf{0.119} & \textbf{0.020} & \textbf{1.530} \\
\cmidrule(l){1-9}
\ & & & & & 0.085 & 0.539 & 0.072 & / \\
\ & \CheckmarkBold & & & & 0.054 & 0.454 & 0.048 & / \\
Dark Road & \CheckmarkBold & \CheckmarkBold & \CheckmarkBold & &0.033 & 0.318 & 0.026 & \textbf{1.278} \\
\ & \CheckmarkBold & \CheckmarkBold &  & \CheckmarkBold &\textbf{0.018} & 0.257 & 0.012 & 1.435 \\
\ & \CheckmarkBold & \CheckmarkBold & \CheckmarkBold & \CheckmarkBold & \textbf{0.015} & \textbf{0.230} & \textbf{0.009} & \textbf{1.303} \\
\bottomrule
\end{tabular}
}
\begin{tablenotes} 
		\item Note: The best and near-best results values for each metric are \\ highlighted in bold.
     \end{tablenotes} 
\label{area_table}
\end{threeparttable}
\end{table}

\begin{figure}[htbp]
\centerline{\includegraphics[width=0.5\textwidth]{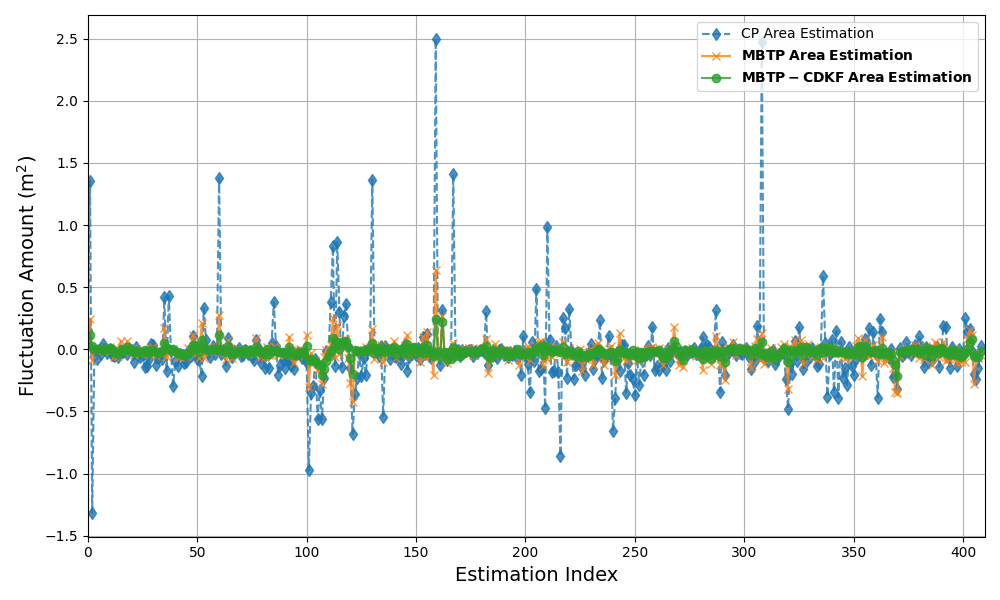}}
\caption{Fluctuation comparison of pothole area estimation using different estimation methods and consecutive frame optimization strategies on the Clear Dataset.}
\label{fluct_show}
\end{figure}

\begin{figure*}[htbp]
\centerline{\includegraphics[width=1.0\textwidth]{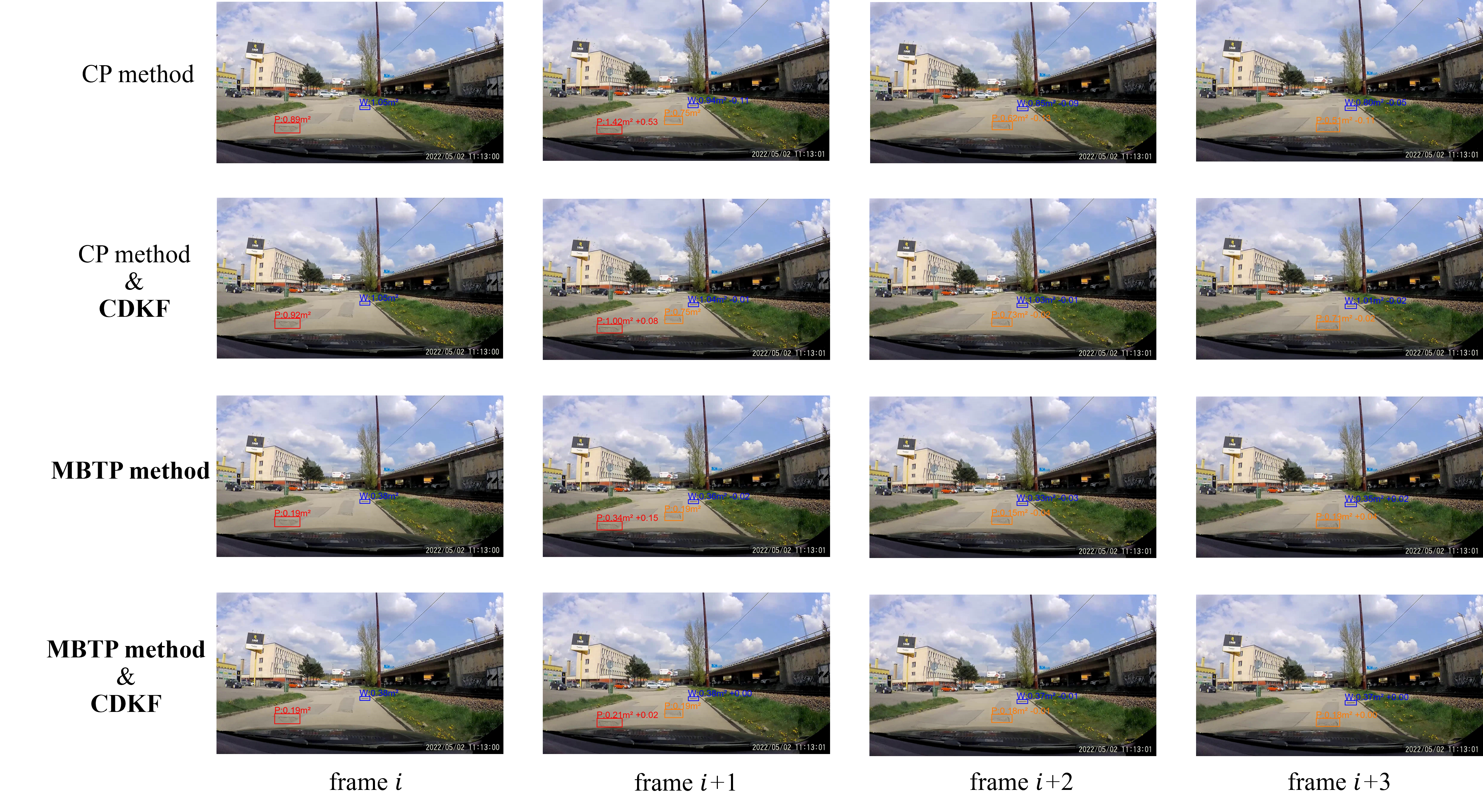}}
\caption{Visualization of area estimation fluctuations for the same pothole across consecutive frames using the different area estimation method and optimization algorithm.}
\label{area_show}
\end{figure*}
The CP method estimates the pothole area by using the depths of two diagonal points of the bounding box, assuming the pothole as a flat rectangular region. However, in real-world scenarios, object detection and depth estimation models introduce errors, particularly on damaged road surfaces, leading to significant inaccuracies. In contrast, our newly proposed MBTP method first maps the pothole to its minimum bounding rectangle, providing a more realistic approximation of its true shape. It then subdivides the pothole into multiple pixel-level triangular facets and incorporates depth information for estimation. This approach minimizes the impact of depth errors from individual pixels, resulting in a more accurate and robust estimation. The superior performance of our method highlights its effectiveness and reliability in pothole area estimation.\par

For the CDKF optimization of pothole estimation across consecutive frames, we compare different measurement noise covariance strategies: using only confidence, using only pothole distance, and combining both with weighted integration, as defined in Eq.~\ref{area_kal_R}. The weights, $\lambda$ and $\theta$, are optimized using Bayesian optimization. For the Clear Road Dataset, $\lambda$ is set to 1.026 and $\theta$ to 0.7179. For the Dark Road Dataset, $\lambda$ is set to 1.51 and $\theta$ to 1.227. The results of these three measurement noise covariance strategies are shown in the lower section of Table.~\ref{area_table}.\par
To validate the effectiveness of the proposed MBTP pothole area estimation algorithm and the consecutive frame optimization strategy CDKF, a series of comparative and ablation experiments were conducted on both datasets, as shown in Table.~\ref{area_table}. The best and near-best results are highlighted in bold. The area estimation methods are evaluated using three metrics: MAE, CV, and AFD.\par
The first two rows of the Table.~\ref{area_table} present the results for the CP method and the proposed MBTP method, respectively. The MBTP method outperforms the CP method across all three metrics, demonstrating its robustness and reliability. The CP method estimates the pothole area by using the depths of two diagonal points within the bounding box and assumes the pothole is a flat rectangle. However, in real-world scenarios, due to inherent errors in both object detection and depth estimation models, particularly under challenging road conditions, this approach often results in significant inaccuracies. In contrast, the MBTP method first maps the pothole to its minimum bounding rectangle, which provides a more realistic representation of the pothole’s shape. It then divides the region into multiple pixel-level triangular facets and integrates depth information to calculate the area. This helps reduce the impact of individual pixel-level depth errors and results in more accurate and robust estimates.\par
The following three rows present the results of applying Kalman filtering to MBTP-based estimates across consecutive video frames. An additional metric, NIS, is introduced to evaluate the consistency and noise modeling performance of the filter.\par
Across both datasets, applying confidence-based or distance-based noise covariance significantly improves MAE, CV, and AFD, indicating enhanced consistency and robustness. When using confidence alone, the NIS values of 1.404 and 1.278 are closest to the ideal value of 1, suggesting more accurate noise modeling. However, this method performs worse on MAE and CV, likely due to the high variability of confidence scores under different scenes. Confidence scores are also inherently non-linear and non-smooth and tend to reflect classification certainty rather than spatial accuracy of the detected regions.\par
Using distance alone produces near-best MAE and CV values, indicating low overall fluctuations and a strong correlation between distance and the reliability of area estimation. However, this approach results in the worst AFD and NIS scores. The high AFD may be due to abrupt changes in the filter’s reliance on measurements when a pothole moves from far to near, leading to inconsistency between consecutive frames. The poor NIS indicates that modeling noise solely with distance is incomplete, likely underestimating the true noise level and resulting in overly large innovations.\par
To address these shortcomings, the combined approach CDKF is proposed. It fuses both confidence and distance through a weighted sum. This combined approach achieves the best AFD scores of 0.02 and 0.009 for the two datasets. It also delivers the best or near-best performance across the other three metrics, achieving a balanced performance overall. For the AFD metric in particular, the use of both factors helps smooth out sudden changes by allowing one factor to compensate when the other varies sharply. This prevents large fluctuations in the Kalman gain and ensures smoother outputs across frames. The MAE and CV results are close to those of the distance-only method, while the NIS values are similar to the confidence-only method.\par

The line chart of area estimation fluctuations for the same pothole using different estimation methods and consecutive frame optimization strategies is shown in Fig.~\ref{fluct_show}. The blue dashed line represents the CP method, which exhibits significant fluctuations and poor stability. The yellow line corresponds to the proposed MBTP method, which shows noticeably reduced variation. The green line represents the MBTP method with CDKF optimization, further minimizing fluctuations and demonstrating improved robustness.\par

The visualization of area estimation results across consecutive frames is shown in Fig.~\ref{area_show}. Each column represents a single video frame, illustrating the area estimation results for four consecutive frames. The four rows correspond to the CP method and its CDKF-optimized version, as well as the MBTP method and its CDKF-optimized counterpart. Bounding boxes of the same color denote the same pothole or well with a consistent ID.\par

The CP method tends to produce overestimated results. As shown in the Fig.~\ref{area_show}, medium-sized potholes are predicted to be approximately 1 m², with significant fluctuations between adjacent frames. For instance, the first two frames exhibit a variation of 0.53 m². This instability may stem from the method's reliance solely on corner features, making it highly sensitive to geometric modeling parameters. Although the CDKF-based optimization reduces prediction volatility, it fails to fully address the systematic overestimation issue.\par

In contrast, the proposed MBTP method yields more reasonable area estimates. The predicted pothole sizes in the figure average around 0.2 m², aligning well with ground truth measurements. While minor fluctuations persist for the same target, the MBTP method demonstrates significantly better stability than the CP approach. Further optimization with CDKF enhances robustness, delivering both precise and consistent predictions for the same pothole across frames. These findings demonstrate that combining both uncertainty measures leads to more stable and reliable area estimations.\par

\subsection{Operation Time}

\begin{table*}[ht]
\centering
\caption{Operation Time of the Proposed Framework and Its Components.}
\renewcommand{\arraystretch}{1.5}
\begin{tabular}{|c||c||c|c|c|c|c|c|}
\hline
\multicolumn{2}{|c||}{Running Time/Frame} & Detection & Tracking & Depth Estimation & \multicolumn{2}{c|}{Area Estimation} & Optimization \\
\hhline{|--||-|-|-|-|-|-|}
Serial & \textbf{Parallel} & \textbf{ACSH-YOLOv8} & BoT-SORT & DepthAnything V2 & MBTP & \textbf{MBTP (JIT)} & \textbf{CDKF} \\
\hhline{|--||-|-|-|-|-|-|}
134.7 ms & \textbf{110.2 ms} & \textbf{23.4 ms} & 1.1 ms & 104 ms & 135 ms & \textbf{6.2 ms} & \textbf{$<$ 0.1 ms} \\
\hline
\end{tabular}
\label{time}
\end{table*}

The runtime of the proposed framework for each frame is measured, as summarized in Table.~\ref{time}. The framework consists of five main steps, each timed separately. Among these components, the detection module (ACSH-YOLOv8), the area estimation module (MBTP), and the consecutive frame optimization module (CDKF) are the methods proposed in this paper. All three exhibit low processing times, with ACSH-YOLOv8 taking 23.4 ms, MBTP taking 6.2 ms, and CDKF requiring less than 0.1 ms per frame. To accelerate the computationally intensive area estimation process, the Numba library is employed for just-in-time (JIT) compilation in Python, reducing the per-frame area estimation time from 135 ms to just 6.2 ms. The most time-consuming component is depth estimation, which requires 104 ms per frame. It is expected that with future advances in monocular depth estimation algorithms, further improvements in overall runtime can be achieved. Both serial and parallel processing schemes are evaluated. Since pothole detection and tracking are independent of depth estimation, they can be executed in parallel. Results show that parallelization reduces the overall processing time by 24.5 ms compared to the serial execution, reaching 110.2 ms for each frame.

\section{Conclusion}
In this paper, a robust pothole area estimation framework for video streams is proposed, which integrates object detection and monocular depth estimation. The estimation is further refined using CDKF for consecutive frame optimization. To address the challenges posed by small potholes and complex edge features, the ACSH-YOLOv8 detection network is proposed with a P2 detection head for small objects and integrating the ACmix attention mechanism into the Neck structure. Then the pre-trained monocular metric depth estimation model is utilized to generate pixel-wise depth maps. This paper proposes MBTP, a novel method for pothole area estimation. Using the pinhole camera model, potholes are mapped to 3D space and enclosed by a minimum bounding rectangle. The area is then calculated by tessellating the pothole into triangles and summing their areas. Finally, leveraging video stream data, the CDKF method is prpposed, which optimally adjusts the estimation by incorporating confidence scores and distance information. Experiments show that our method significantly improves detection accuracy, especially for small potholes and complex edges. For area estimation, the MBTP method and CDKF yield more reliable and robust results.\par
The proposed fully vision-based pothole area estimation framework offers an efficient and reliable solution for enhancing the safety and comfort of autonomous driving. However, certain limitations remain. The method struggles with detecting highly blurred potholes, and modeling noise solely based on confidence and distance may not fully capture real-world variations. Additionally, the overall pipeline lacks dedicated runtime optimization to reduce latency. In future work, we plan to further refine the pothole detection network, incorporate factors such as ambient lighting and vehicle stability into noise modeling, and explore shared backbone architectures or parallel optimization techniques to improve area estimation speed, enhancing the framework’s practical viability.\par

\bibliographystyle{unsrt}
\bibliography{reference}

\end{document}